\documentclass{article}

\usepackage{microtype}
\usepackage{graphicx}
\usepackage{subfigure}
\usepackage{booktabs} 
\usepackage[algo2e]{algorithm2e}

\usepackage{hyperref}

\usepackage[arxiv]{icml2025}

\usepackage{amsmath}
\usepackage{amssymb}
\usepackage{mathtools}
\usepackage{amsthm}

\usepackage[capitalize,noabbrev]{cleveref}

\theoremstyle{plain}

\theoremstyle{definition}

\theoremstyle{remark}

\usepackage[textsize=tiny]{todonotes}

\icmltitlerunning{With Great Backbones Comes Great Adversarial Transferability}

\begin{document}

\twocolumn[
\icmltitle{With Great Backbones Comes Great Adversarial Transferability}



\icmlsetsymbol{equal}{$\dagger$}

\begin{icmlauthorlist}
\icmlauthor{Erik Arakelyan\texorpdfstring{$^\dagger$}{+}}{denmark}
\icmlauthor{Karen Hambardzumyan}{equal,yerevann,ucl}
\icmlauthor{Davit Papikyan}{yerevann}
\icmlauthor{Pasquale Minervini}{edinburgh}
\icmlauthor{Albert Gordo}{personal}
\icmlauthor{Isabelle Augenstein}{denmark}
\icmlauthor{Aram H. Markosyan}{yerevann}
\end{icmlauthorlist}

\icmlaffiliation{denmark}{Department of Computer Science, University Of Copenhagen, Denmark}
\icmlaffiliation{yerevann}{YerevaNN, Armenia}
\icmlaffiliation{personal}{Independent Researcher}
\icmlaffiliation{edinburgh}{School of Informatics, University Of Edinburgh, United Kingdom}
\icmlaffiliation{ucl}{UCL Centre for Artificial Intelligence, University College London}

\icmlcorrespondingauthor{Erik Arakelyan}{erik.a@di.ku.dk}

\icmlkeywords{Machine Learning, Adversarial Attacks, Deep Learning, Adversarial Robustness, Adversarial Transferability}

\vskip 0.3in
]



\printAffiliationsAndNotice{\icmlEqualContribution} 

\begin{abstract}
Advancements in self-supervised learning (SSL) for machine vision have enhanced representation robustness and model performance, leading to the emergence of publicly shared pre-trained backbones, such as \emph{ResNet} and \emph{ViT} models tuned with SSL methods like \emph{SimCLR}. Due to the computational and data demands of pre-training, the utilization of such backbones becomes a strenuous necessity. However, employing such backbones may imply adhering to the existing vulnerabilities towards adversarial attacks.
Prior research on adversarial robustness typically examines attacks with either full (\emph{white-box}) or no access (\emph{black-box}) to the target model, but the adversarial robustness of models tuned on known pre-trained backbones remains largely unexplored. Furthermore, it is unclear which tuning meta-information is critical for mitigating exploitation risks. In this work, we systematically study the adversarial robustness of models that use such backbones, evaluating $20000$ combinations of tuning meta-information, including fine-tuning techniques, backbone families, datasets, and attack types. 
 To uncover and exploit potential vulnerabilities, we propose using proxy (surrogate) models to transfer adversarial attacks, fine-tuning these proxies with various tuning variations to simulate different levels of knowledge about the target. Our findings show that proxy-based attacks can reach close performance to strong \emph{black-box} methods with sizable budgets and closing to \emph{white-box} methods, exposing vulnerabilities even with minimal tuning knowledge. Additionally, we introduce a naive "backbone attack", leveraging only the shared backbone to create adversarial samples, demonstrating an efficacy surpassing \emph{black-box} and close to \emph{white-box} attacks and exposing critical risks in model-sharing practices. Finally, our ablations reveal how increasing tuning meta-information impacts attack transferability, measuring each meta-information combination.
\end{abstract}

\section{Introduction}

Machine vision models pre-trained with massive amounts of data and using self-supervised techniques \citep{newell2020useful} are shown to be robust and highly performing\citep{goyal2021self,goldblum2024battle} feature-extracting backbones \citep{elharrouss2022backbones,han2022survey}, which are further used in a variety of tasks, from classification \citep{atito2021sit,chen2020big} to semantic segmentation \citep{ziegler2022self}. However, creating such backbones incurs substantial data annotation \citep{jing2020self} and computational costs \citep{han2022survey}, consequently rendering the use of such publicly available pre-trained backbones the most common and efficient solution for researchers and engineers alike. 
Prior works have focused on analysing safety and adversarial robustness with complete, i.e. \emph{white-box} \citep{porkodi2018survey} or no, i.e. \emph{black-box} \citep{bhambri2019survey} knowledge of the target model weights, fine-tuning data, fine-tuning techniques and other tuning meta-information. Although, in practice, an attacker can access partial knowledge \citep{DBLP:conf/iclr/LordMB22,DBLP:journals/tip/ZhuCLC00ZCH22,DBLP:conf/sp/CarliniCN0TT22} of how the targeted model was produced, i.e. original backbone weights, tuning recipe, etc., the adversarial robustness of models tuned on a downstream task from a given pre-trained backbone remains largely underexplored. We refer to settings with partial knowledge of target model constructions meta-information as \emph{grey-box}. This is important both for research and production settings because with an increased usage \citep{DBLP:conf/nips/GoldblumSNSPSCI23} of publically available pre-trained backbones for downstream applications, we are incapable of assessing the potential exploitation susceptibility and inherent risks within models tuned on top of them and subsequently enhance future pre-trained backbone sharing practices.
\begin{figure*}[t!]
    \centering
    \includegraphics[clip=true,width=\textwidth]{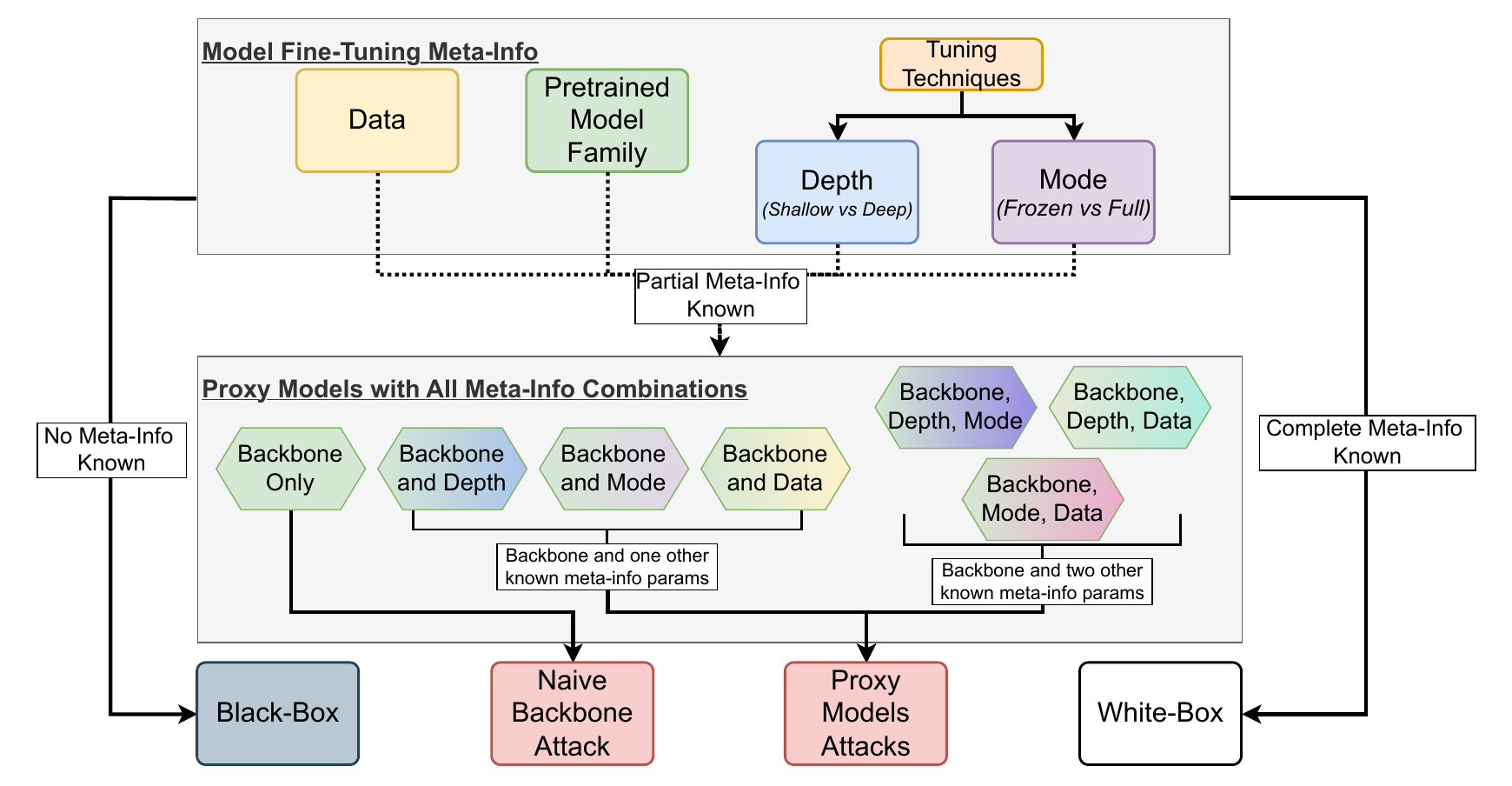}
    \caption{The figure depicts all of the settings used to evaluate adversarial vulnerabilities given different information of the target model construction. From left to right, we simulate exhaustive varying combinations of meta-information available about the target model during adversarial attack construction. All of the created proxy models are used separately to assess adversarial transferability.}
    \label{fig:framework}
\end{figure*}

In this work, we systematically explore the safety towards adversarial attacks within the models tuned on a downstream classification task from a known publically available backbone pre-trained with a self-supervised objective. We further explicitly measure the effect of the target model construction meta-information by simulating different levels of its availability during the adversarial attack. 
For this purpose, we initially train $352$ diverse models from $21$ families of commonly used pre-trained backbones using $4$ different fine-tuning techniques and $4$ datasets. We fix each of these networks as a potential target model and transfer adversarial attacks using all of the other models produced from the same backbones as proxy surrogates \citep{DBLP:conf/aaai/QinXYH23,DBLP:conf/iclr/LordMB22} for adversarial attack construction. Each surrogate model simulates varying levels of knowledge availability w.r.t. target model construction on top of the available backbone during adversarial attack construction. This constitutes approximately $20000$ adversarial transferability comparisons between target and proxy pairs across all model families and meta-information variations.
By assessing the adversarial transferability of attacks from these surrogate models, we are able to explicitly measure the impact of the availability of each meta-information combination about the final target model during adversarial sample generation.

We further introduce a naive exploitation method referred to as \emph{backbone attacks} that utilizes only the pre-trained feature extractor for adversarial sample construction. The attack uses projected gradient descent over the representation space to disentangle the features of similar examples. Our results show that both proxy models and even simplistic \emph{backbone attacks} are capable of surpassing strong query-based \emph{black-box} methods and closing to \emph{white-box} performance. The findings indicate that \emph{backbone attacks}, where the attacker lacks meta-information about the target model, are generally more effective than attempts to generate adversarial samples with limited knowledge. This highlights the vulnerability of models built on publicly available backbones.

Our ablations show that \textit{having access to the weights of the pre-trained backbone is functionally equivalent to possessing all other meta-information about the target model when performing adversarial attacks}. We compare these two scenarios and show that both lead to similar vulnerabilities, highlighting the interchangeable nature of these knowledge types in attack effectiveness. Our results emphasize the risks in sharing and deploying pre-trained backbones, particularly concerning the disclosure of meta-information. Our experimental framework can be seen in \cref{fig:framework}.

Toward this end, our contributions are as follows:

\begin{itemize} 
\item We introduce, formalize and systematically study the \textbf{grey-box} adversarial setting, which reflects realistic scenarios where attackers have partial knowledge of target model construction, such as access to pre-trained backbone weights and/or fine-tuning meta-information. 
\item We simulate over $20,000$ adversarial transferability comparisons, evaluating the impact of varying levels of meta-information availability about target models during attack construction.
\item We propose a naive attack method, \emph{backbone attacks}, which leverages the pre-trained backbone's representation space for adversarial sample generation, demonstrating that even such a simplistic approach can achieve stronger performance compared to a query-based black-box method and often approaches white-box attack effectiveness. 
\item We show that access to pre-trained backbone weights alone enables adversarial attacks as effectively as access to the full meta-information about the target model, emphasizing the inherent vulnerabilities in publicly available pre-trained backbones.  \end{itemize}

\section{Related Work}

\paragraph{Self Supervised Learning}

With the emergence of massive unannotated datasets in machine vision, such as YFCC100M\citep{DBLP:journals/cacm/ThomeeSFENPBL16}, ImageNet\citep{deng2009imagenet}, CIFAR \citep{krizhevsky2009learning} and others Self Supervised Learning (SSL) techniques \citep{DBLP:journals/pami/JingT21} became increasingly more popular for pre-training the models \citep{newell2020useful}. This prompted the creation of various families of SSL objectives, such as colorization prediction \citep{DBLP:conf/eccv/ZhangIE16}, jigsaw puzzle solving \citep{DBLP:conf/eccv/NorooziF16} with further invariance constraints \citep[PIRL]{DBLP:conf/cvpr/MisraM20}, non-parametric instance discrimination \citep[NPID, NPID++]{DBLP:conf/cvpr/WuXYL18}, unsupervised clustering \citep{DBLP:conf/eccv/CaronBJD18}, rotation prediction \citep[RotNet]{DBLP:conf/iclr/GidarisSK18}, sample clustering with cluster assignment constraints\citep[SwAV]{DBLP:conf/nips/CaronMMGBJ20},  contrastive representation entanglement \citep[SimCLR]{DBLP:conf/icml/ChenK0H20},
self-distillation without labels \citep[DINO]{DBLP:conf/iccv/CaronTMJMBJ21} and others \citep{DBLP:journals/pami/JingT21}. Numerous architectures, like AlexNet \citep{DBLP:conf/nips/KrizhevskySH12}, variants of ResNet\citep{DBLP:conf/cvpr/HeZRS16} and visual transformers \citep{DBLP:conf/iclr/DosovitskiyB0WZ21,DBLP:conf/icml/TouvronCDMSJ21,DBLP:conf/nips/AliTCBDJLNSVJ21} were trained using these SSL methods and shared for public use, thus forming the set of widely used pre-trained backbones.
We obtain all of these models trained with different self-supervised objectives from their original designated studies summarised in VISSL \citep{goyal2021vissl}. An exhaustive list of all models can be seen in \cref{tab:ssl_model_summary}.

\begin{table}[!t]
\centering
\resizebox{\columnwidth}{!}{%
\begin{tabular}{@{}lcc@{}}
\toprule
\textbf{SSL Method} & \textbf{Pretraining Dataset} & \textbf{Architecture} \\ 
\midrule
\multicolumn{3}{l}{\textbf{Colorization} \citep{DBLP:conf/eccv/ZhangIE16}} \\
Colorization & YFCC100M      & AlexNet   \\
Colorization & ImageNet-1K   & AlexNet   \\
Colorization & ImageNet-1K   & ResNet-50 \\
Colorization & ImageNet-21K  & AlexNet   \\
Colorization & ImageNet-21K  & ResNet-50 \\
\midrule
\multicolumn{3}{l}{\textbf{Jigsaw Puzzle}\citep{DBLP:conf/eccv/NorooziF16}} \\
Jigsaw Puzzle & ImageNet-21K & ResNet-50  \\
Jigsaw Puzzle & ImageNet-1K  & ResNet-50  \\
Jigsaw Puzzle & ImageNet-21K & ResNet-50  \\
Jigsaw Puzzle & ImageNet-21K & AlexNet    \\
Jigsaw Puzzle & ImageNet-1K  & AlexNet    \\
Jigsaw Puzzle & ImageNet-1K  & ResNet-50  \\
\midrule
\multicolumn{3}{l}{\textbf{PIRL (Jigsaw-based)}\citep{DBLP:conf/cvpr/MisraM20}} \\
PIRL         & ImageNet-1K   & ResNet-50  \\
\midrule
\multicolumn{3}{l}{\textbf{Rotation Prediction} \citep{DBLP:conf/iclr/GidarisSK18}} \\
RotNet       & ImageNet-1K   & ResNet-50  \\
\midrule
\multicolumn{3}{l}{\textbf{DINO}\citep{DBLP:conf/iccv/CaronTMJMBJ21}} \\
DINO         & ImageNet-1K   & DeiT-Small \\
DINO         & ImageNet-1K   & XCiT-Small \\
\midrule
\multicolumn{3}{l}{\textbf{SimCLR}\citep{DBLP:conf/icml/ChenK0H20}} \\
SimCLR       & ImageNet-1K   & ResNet-50  \\
SimCLR       & ImageNet-1K   & ResNet-101 \\
\midrule
\multicolumn{3}{l}{\textbf{SwAV} \citep{DBLP:conf/nips/CaronMMGBJ20}} \\
SwAV         & ImageNet-1K   & ResNet-50  \\
SwAV         & ImageNet-1K   & ResNet-50  \\
\midrule
\multicolumn{3}{l}{\textbf{DeepCluster V2} \citep{DBLP:conf/eccv/CaronBJD18}} \\
DeepCluster V2 & ImageNet-1K & AlexNet    \\
\midrule
\multicolumn{3}{l}{\textbf{Instance Discrimination (NPID)} \citep{DBLP:conf/cvpr/WuXYL18}} \\
NPID         & ImageNet-1K   & ResNet-50  \\
\bottomrule
\end{tabular}
}
\caption{Summary of Self-Supervised Learning Methods, Pretraining Datasets, and Architectures used in our study.}
\label{tab:ssl_model_summary}
\end{table}

\paragraph{Adversarial Attacks}
The availability of pre-trained backbones allows to test them for vulnerabilities towards adversarial attacks, which are learnable imperceptible perturbations generated to mislead models into making incorrect predictions \citep{DBLP:journals/corr/SzegedyZSBEGF13, DBLP:journals/corr/GoodfellowSS14}. Several attack strategies have been studied, including single-step fast gradient descent \citep[FGSM]{DBLP:journals/corr/GoodfellowPMXWOCB14,DBLP:conf/iclr/KurakinGB17a}, and computationally more expensive optimization-based attacks, such as projected gradient descent based attacks \citep[PGD]{DBLP:conf/iclr/MadryMSTV18}, CW \citep{DBLP:conf/ccs/Carlini017}, JSMA \citep{DBLP:conf/ccs/PapernotMGJCS17}, and others \citep{DBLP:conf/cvpr/DongLPS0HL18, DBLP:conf/cvpr/Moosavi-Dezfooli16,DBLP:conf/iclr/MadryMSTV18}. All of these attacks assume complete access to the target model, which is known as the \emph{white-box} \citep{DBLP:conf/ccs/PapernotMGJCS17} setting. These attacks can be \emph{targeted} toward confusing the model to infer a specific wrong class or \emph{untargeted} with the desire that it infers any incorrect label. However, an opposite setting with no information, referred to as \emph{black-box} \citep{DBLP:conf/ccs/PapernotMGJCS17}, has also been explored as a more practical setting. The methods involve attempts at gradient estimation \citep{DBLP:conf/ccs/ChenZSYH17,DBLP:conf/icml/IlyasEAL18,DBLP:conf/eccv/BhagojiHLS18}, adversarial transferability \citep{DBLP:conf/ccs/PapernotMGJCS17, DBLP:conf/eccv/ChenZHW20}, local search \citep{DBLP:journals/corr/NarodytskaK16, DBLP:conf/iclr/BrendelRB18,DBLP:conf/icml/LiLWZG19,DBLP:conf/icml/MoonAS19}, combinatorial perturbations \citep{DBLP:conf/icml/MoonAS19} and others \citep{bhambri2019survey}. However, these methods also require massive sample query budgets ranging from $\left[10^3, 10^5 \right]$ queries or computational resources creating each adversarial sample \citep{bhambri2019survey}. Compared to these, we introduce a novel setup with the knowledge of the pre-trained backbone and varying levels of partially known target model tuning meta-information during adversarial attack construction, which we call \emph{grey-box}. We show that even simple naive attacks are capable of exploiting better than black-box attacks without the need for significantly querying the target model.

\paragraph{Adversarial Transferability}

Our work is also aligned with adversarial transferability, where adversarial examples generated for one model can mislead other models, even without access to the target model weights or training data. This property poses significant security concerns, as it allows for effective black-box attacks on systems with no direct access \citep{DBLP:conf/ccs/PapernotMGJCS17, DBLP:conf/icml/IlyasEAL18}. Efforts can be divided into \emph{generation-based} and \emph{optimisation} methods. Generative methods have emerged as an alternative approach to iterative attacks, where adversarial generators are trained to produce transferable perturbations. For instance, \citet{DBLP:conf/cvpr/PoursaeedKGB18} employed autoencoders trained on white-box models to generate adversarial examples. Most of the attacks aiming for adversarial transferability strongly depend on the availability of data from the target domain \citep{DBLP:conf/ccs/Carlini017, DBLP:conf/ccs/PapernotMGJCS17}. However, although current adversarial transferability methods claim to produce massive vulnerabilities in machine vision models, \citet{DBLP:journals/corr/abs-2105-00433} examines the practical implications of adversarial transferability, which are frequently overstated. That study demonstrates that it is nearly impossible to reliably predict whether a specific adversarial example will transfer to an unseen target model in a black-box setting.  This perspective underscores the importance of systematically evaluating transferability in realistic settings, including scenarios where attackers are sensitive to the cost of failed attempts.
In our study, we offer a novel systematic approach to explicitly assess the adversarial transferability with varying levels of meta-information knowledge.

\section{Methodology}

\paragraph{Preliminaries}
For consistency, we employ the following notation. We denote each Dataset $\mathcal{D}=\{\mathcal{X},\mathcal{Y}\}$. Where $\mathcal{X}=\{x_1, \dots ,x_{|\mathcal{D}|}\}$ is a set of images, with $x_i \in \mathcal{R}^{H \times W \times C}$, where $H$,$W$ and $C$ are the height, width and the channels of the image accordingly and $\mathcal{Y} = \{y_1 \dots y_n\}$ is used as the set of ground truth labels. We denote the training, validation and testing splits per task as $\mathcal{D}=\{\mathcal{D}_{train}, \mathcal{D}_{val}, \mathcal{D}_{test}\}$. 
A \textit{model} is defined as the following tuple $\mathcal{M} = \mathcal{M}(\mathcal{D}, \mathcal{W}, \mathcal{B}, \mathcal{F})$, where $\mathcal{D}$ contains the dataset used for training, $\mathcal{W}$ are the weights of the trained model and $\mathcal{B}$ is the pre-trained back-bone $\mathcal{B}( \mathcal{W}_{\mathcal{B}})$ with available weights $\mathcal{W}_{\mathcal{B}}$. The notation $\mathcal{F}(\mathcal{T}, \mathcal{Z})$, where $\mathcal{T}$ encodes the \emph{mode} of tuning (e.g., full fine-tuning, partial fine-tuning, etc.) and $\mathcal{Z}$ the \emph{depth} of tuning of the final classifier on top of the backbone.

\paragraph{Meta-Information variations}
We define the variations of the available meta-information about the target model $\mathcal{M}$ during an adversarial attack as a \textit{unit of release} $\mathcal{R}=\mathcal{R}(\mathcal{M}(\mathcal{D}, \mathcal{W}, \mathcal{B}(\mathcal{W}_{\mathcal{B}}), \mathcal{F}(\mathcal{T}, \mathcal{Z})))$. For example, if the target fine-tuning mode $\mathcal{Z}^\textrm{target}$ and dataset $\mathcal{D}^\textrm{target}$ are not known, the unit of release will be $\mathcal{R}=\mathcal{R}(\mathcal{M}(*, \mathcal{W}, \mathcal{B}(\mathcal{W}_{\mathcal{B}}), \mathcal{F}(\mathcal{T}, *)))$. Note that the \emph{black-box} setting will correspond to the unit of release $\mathcal{R}(\mathcal{M}(*, *, *, *, *))$ and the \emph{white-box} setting to $\mathcal{R}(\mathcal{M}(\mathcal{D}, \mathcal{W}, \mathcal{B}(\mathcal{W}_{\mathcal{B}}), \mathcal{F}(\mathcal{T}, \mathcal{Z})))$, all the variations between these are considered \emph{grey-box}. When discussing any experiments within the \emph{gery-box} setup, we assume the minimal unit of release contains knowledge about at least the pre-trained backbone i.e. $\mathcal{R}(\mathcal{M}(*, *, \mathcal{B}(\mathcal{W}_{\mathcal{B}}), *)$.

\paragraph{Adversarial Attacks with Proxy Models}

To test the adversarial robustness of the models trained from the same pre-trained backbone, we create a set of proxy models $\mathcal{M}^\textrm{proxy} = \{\mathcal{M}^\textrm{proxy}_{1} \dots \mathcal{M}^\textrm{proxy}_{v}\}$ given the pre-trained backbone $\mathcal{B}$, where $v$ is the number of all possible units of release between \emph{black-box} and \emph{white-box} settings that include the backbone. For each proxy model $\mathcal{M}^\textrm{proxy}_{i}$ with its designated meta-information unit of release $\mathcal{R}_i$, we use an adversarial attack $\mathcal{A}$ to generate adversarial noise and further transfer it to the target model $\mathcal{M}^\textrm{target}$. This means that given an example image $x$ with a label $y$, target and proxy models $\mathcal{M}^{\textrm{target}}$, $\mathcal{M}^{\textrm{proxy}}$ we want to produce a sample $x'$ that would fool the target model, such that $\arg \max \mathcal{M}^{\textrm{target}} (x') \neq y$. If we are using a targeted attack then we want $\mathcal{M}^{\textrm{target}} (x') = t$ where $t$ is the targeted class different from the ground truth $t \neq c_{gt}$.
After creating the adversarial attack for each sample in $\mathcal{D}_\mathrm{test}^\mathrm{proxy}$ and $\mathcal{D}_\mathrm{test}^\mathrm{target}$ we evaluate the success rate of the attack and the success rate of the transferability onto the target model. To measure the success and robustness of the adversarial attack and its transferability, we define the following metrics:

\begin{itemize}
    \item \textbf{Attack Success Rate (ASR):} This is the proportion of adversarial examples successfully fooling the proxy model $\mathcal{M}_i^{\mathrm{proxy}}$, defined as:
    \begin{equation}
        \text{ASR}_i = \frac{1}{|\mathcal{D}_\mathrm{test}^\mathrm{proxy}|} \sum_{x \in \mathcal{D}_\mathrm{test}^\mathrm{proxy}} \mathbb{I} \left[ \arg \max \mathcal{M}^{\mathrm{proxy}}_i(x') \neq y \right],
    \end{equation}
    where $\mathbb{I}[\cdot]$ is the indicator function.
    \item \textbf{Transfer Success Rate (TSR):} To evaluate the transferability of adversarial examples generated using the proxy model $\mathcal{M}_i^{\mathrm{proxy}}$ to the target model $\mathcal{M}^{\mathrm{target}}$, we compute the fooling rate on the target model as:
    \begin{equation}
        \text{TSR}_i = \frac{1}{|\mathcal{D}_\mathrm{test}^\mathrm{target}|} \sum_{x \in \mathcal{D}_\mathrm{test}^\mathrm{target}} \mathbb{I} \left[ \arg \max \mathcal{M}^{\mathrm{target}}(x') \neq y \right].
    \end{equation}
\end{itemize}

This setup allows us to explicitly quantify how the availability of diverse meta-information combinations explicitly impacts the adversarial transferability of the given model, thus highlighting the risks in the model-sharing practices. A visual depiction of this can be seen in \cref{fig:framework}.
  
\subsection{Backbone Attack}

\begin{algorithm}[!t]
\caption{Backbone Attack}
\label{alg:backbone_pgd}
\KwIn{Model backbone $\mathcal{B}$, clean image $x_0$, perturbation bound $\epsilon$, step size $\alpha$, number of steps $T$, distance function $\mathcal{L}_{\text{cosine}}$, random start flag}
\KwOut{Adversarial image $x_{\text{adv}}$}

\textbf{Initialization:} \\
$x_{\text{adv}} \gets x_0$ 

\If{random start}{
$x_{\text{adv}} \gets x_{\text{adv}} + \text{Uniform}(-\epsilon, \epsilon)$\\
$x_{\text{adv}} \gets \text{Clip}(x_{\text{adv}}, 0, 1)$
}

\textbf{Fixed Original Image Representation:} \\
$z_0 \gets StopGrad(\mathcal{B}(x_0))$\\

\For{$t = 1$ \textbf{to} $T$}{
    \textbf{Forward Pass:} \\
    $z_{\text{adv}} \gets \mathcal{B}(x_{\text{adv}})$ \tcp{Adversarial image representation}
    
    \textbf{Compute Loss and Gradient:} \\
    $\mathcal{L} \gets 1 - \text{cos}(z_{\text{adv}}, z_0)$ \tcp{Distance loss}    
    $g \gets \nabla_{x_{\text{adv}}} \mathcal{L}$ \tcp{Gradient w.r.t $x_{\text{adv}}$}
    
    \textbf{Update Adversarial Image:} \\
    $x_{\text{adv}} \gets x_{\text{adv}} + \alpha \cdot \text{sign}(g)$ \tcp{PGD step}
    
    \textbf{Projection:} \\
    $\delta \gets \text{Clip}(x_{\text{adv}} - x_0, -\epsilon, \epsilon)$ \tcp{Project perturbation into $\ell_\infty$-ball}
    $x_{\text{adv}} \gets \text{Clip}(x_0 + \delta, 0, 1)$ \tcp{pixel range}
}

\Return $x_{\text{adv}}$
\end{algorithm}

To test the vulnerabilities associated with publicly available pre-trained feature extractors, we designed a naive \emph{backbone attack}, which only utilises the known backbone $\mathcal{B}$ of the model $\mathcal{M}^{\textrm{target}}$. The aim, similar to the prior paragraph, is to create an adversarial attack from the $\mathcal{B}$ to transfer towards the target model $\mathcal{M}^{\mathrm{target}}$. To do this, we utilise a Projected Gradient Descent \citep[PGD]{DBLP:conf/iclr/MadryMSTV18}-based method, where the attack iteratively perturbs the input images in order to maximise the distance between the feature representations of the clean input and the adversarial input, as derived from the backbone $\mathcal{B}$. More formally, let $x$ and $\tilde{x}$ represent the clean input and adversarial input, respectively. The attack iteratively refines $\tilde{x}$ such that:
\begin{align}
    \tilde{x}_{t+1} = \text{Proj}_{\mathcal{S}} \left( \tilde{x}_t + \alpha \cdot \text{sign}\left( \nabla_{\tilde{x}_t} \mathcal{L}_{\mathcal{B}}(x, \tilde{x}_t) \right) \right),
\end{align}
where $\mathcal{L}_{\mathcal{B}}$ is the loss function defined to measure the distance between the feature representations of the clean and adversarial inputs. The backbone representations $f_{\mathcal{B}}$ are extracted as $f_{\mathcal{B}}(x) = \mathcal{B}(x)$, and the differentiable loss can be formulated as:
\begin{align}
    \mathcal{L}_{\mathcal{B}}(x, \tilde{x}) = 1 - \text{cos} \left( f_{\mathcal{B}}(x), f_{\mathcal{B}}(\tilde{x}) \right),
\end{align}
where $\text{cos}(\cdot, \cdot)$ represents the cosine similarity between the two feature vectors. To prevent gradient computation from propagating to the clean representation $f_{\mathcal{B}}(x)$, we utilize a stop-gradient operation $\tilde{f}_{\mathcal{B}}(x) = \textrm{SG}(f_{\mathcal{B}}(x))$.
The adversarial input $\tilde{x}$ is initialized with a random perturbation within the $\ell_{\infty}$ ball of radius $\epsilon$, and the updates are iteratively projected back onto this ball using the $\text{Proj}_{\mathcal{S}}$ operator:
\begin{align}
    \text{Proj}_{\mathcal{S}}(\tilde{x}) = \text{clip} \left( x + \delta, 0, 1 \right), \quad \\ \nonumber\text{where} \quad \delta = \text{clip} \left( \tilde{x} - x, -\epsilon, \epsilon \right).
\end{align}

The pseudo-code of the complete process can bee seen in \cref{alg:backbone_pgd}. In summary, the backbone attack focuses solely on the backbone $\mathcal{B}$, without requiring any knowledge of the full target model $\mathcal{M}^{\textrm{target}}$, thereby revealing vulnerabilities inherent to publicly available feature extractors.

\section{Experimental Setup}

\begin{figure*}[t!]
    \centering
    \includegraphics[clip=true,width=\textwidth]{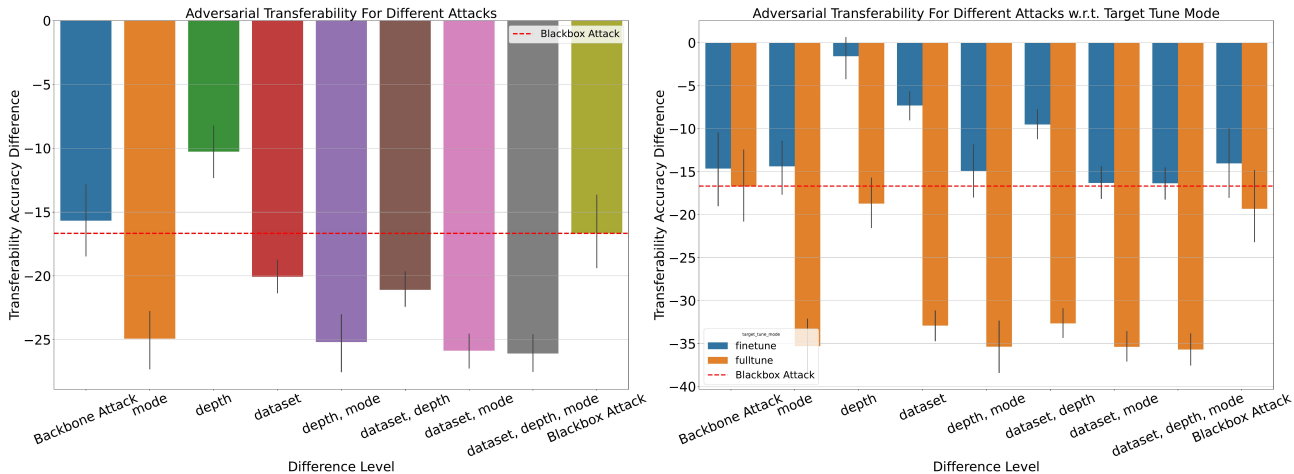}
    \caption{The figure depicts the impact of the \textbf{unavailability}, i.e. difference from the target model, with each possible meta-information combination on adversarial transferability during proxy attack construction and the backbone attack. The results show the average difference from the \emph{white-box} in transferability using PGD with a higher budget (left) and the segmentation w.r.t. in the target training mode (right).}
    \label{fig:whitebox_diff_per_levels}
\end{figure*}

\begin{figure}[t!]
    \centering
    \includegraphics[clip=true,width=\columnwidth]{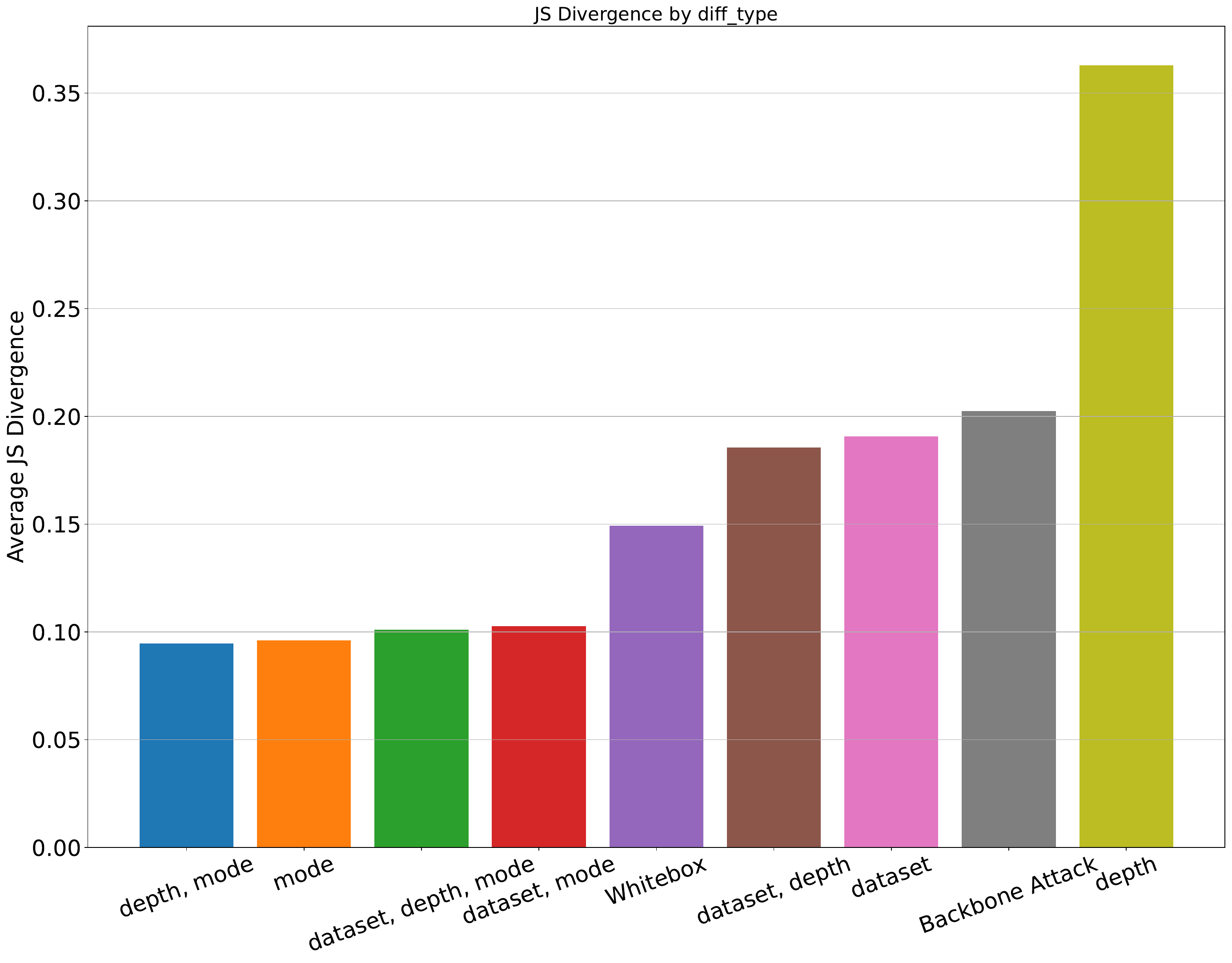}
    \caption{The figure breaks down impact of the \textbf{unavailability}, i.e. difference from the target model, of each possible meta-information combination on the change in the final decision-making of the model. Higher JS divergence implies a bigger change in the final classification of the sample.}
    \label{fig:whitebox_diff_per_levels_tuning}
\end{figure}

\paragraph{Image classification datasets}

Through our study, we use $4$ datasets covering both classical and domain-specific classification benchmarks, such as CIFAR-10 and CIFAR-100 \citep{beyer2020we} and  Oxford-IIIT Pets \citep{parkhi2012cats}, Oxford Flowers-102 \citep{nilsback2008automated}. We train the proxy and target model variation on each one of the datasets using the recipe from \citep{kolesnikov2020big}, reproducing the state-of-the-art model performance results \citep{dosovitskiy2020image, yu2022coca,bruno2022efficient,foret2020sharpness}.

\paragraph{Model variations}
We use $21$ different models tuned from $5$ architectures, $9$ self-supervised objectives and $3$ pre-training datasets. A detailed overview of these can be seen in \cref{tab:ssl_model_summary}.

\paragraph{Model Fintuning Variations}
For training the proxy and target models, we employ two \emph{modes} of training $\mathcal{T}$, with full-tuning of the weights and with fine-tuning only the last added classification layers on top of the pre-trained backbone. We also define the depth of tuning $\mathcal{Z}$ as the number of classification layers added on top of the pre-trained backbone. We use $\{1, 3\}$ final layers corresponding to \emph{shallow} and \emph{deep} tuning settings.

\paragraph{Adversarial Attacks}
To assess the \emph{white-box} adversarial attack success rate and the adversarial transferability from the proxy models, we employ FGSM \citep{DBLP:journals/corr/GoodfellowSS14} and PGD \citep{DBLP:conf/iclr/MadryMSTV18}. We use standard attack hyper-parameters introduced in parallel adversarial transferability studies \citep{DBLP:conf/wacv/WasedaNLNE23, DBLP:conf/iclr/NaseerR0KP22}. For a fair comparison, we also use the same values for our \emph{backbone-attack}.
To show that our results are consistent even with a higher computational budget, we report the results of PGD with $4$ times more iterations per sample for \emph{white-box}, proxy and \emph{backbone} attack experiments.
For \emph{black-box} experiments, we use the Square attack \citep{DBLP:conf/eccv/AndriushchenkoC20}, which is a query-efficient method that uses a random search through adversarial sample construction. To standardise the query budget for all architectures and simulate real-world constraints, we allow $10$ queries of the target model per sample.

\begin{figure*}[t!]
    \centering
    \includegraphics[clip=true,width=\textwidth]{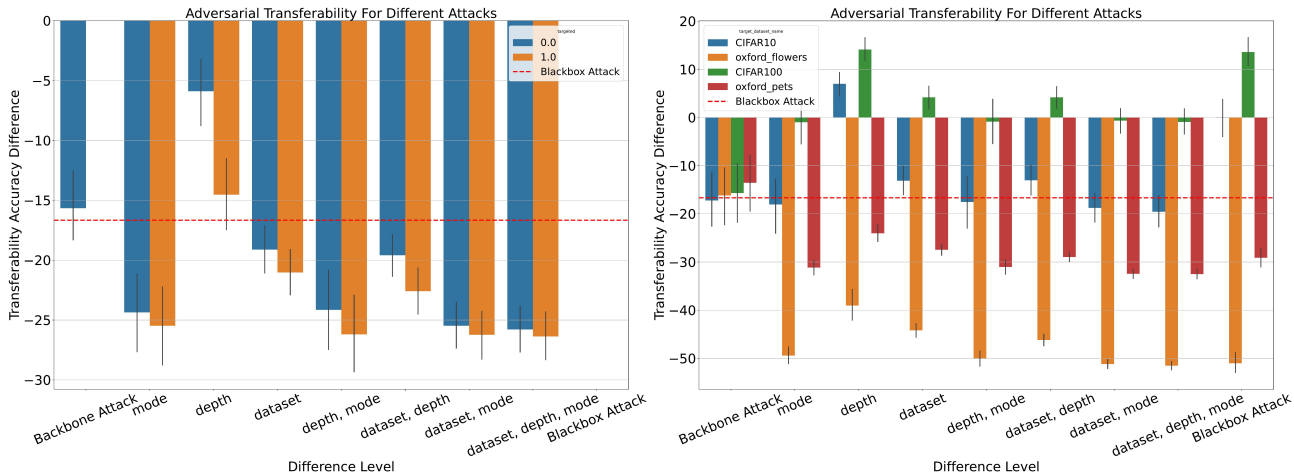}
    \caption{The figure depicts the impact of the \textbf{unavailability}, i.e. difference from the target model, of each possible meta-information combination on adversarial transferability during proxy attack construction and the backbone attack. The results show the average transferability for PGD with a higher budget for targeted vs untargeted attacks (left) and the segmentation w.r.t. the target training dataset (right).}
    \label{fig:whitebox_diff_per_levels_data_target}
\end{figure*}
\begin{table}[t!]
\centering
\resizebox{\columnwidth}{!}{%
\begin{tabular}{@{}lcccc@{}}
\toprule
                       & \multicolumn{2}{c}{\textbf{Original Entropy}} & \multicolumn{2}{c}{\textbf{Adversarial Entropy}} \\ \midrule
\textbf{Metadata type} & \textbf{F-Statistic}    & \textbf{P-Value}    & \textbf{F-Statistic} & \textbf{P-Value} \\
\textit{Target Tune Mode}  & 0.00    & 0.96 & 1238.7  & 0.0 \\
\textit{Proxy Tune Mode}   & 0.02    & 0.88 & 0.5     & 0.4 \\
\textit{Target Dataset}    & 2812.25 & 0.00 & 1184.1  & 0.0 \\
\textit{Proxy Dataset}     & 8.31    & 0.00 & 5.0     & 0.0 \\
\textit{Target Tune Depth} & 5.64    & 0.01 & 0.36    & 0   \\
\textit{Proxy Tune Depth}  & 0.08    & 0.77 & 0.00    & 0   \\ \bottomrule
\end{tabular}%
}
\caption{Variance analysis of entropy values across categorical variables. The table shows F-statistics and p-values for both original and adversarial entropy means. Significant p-values (p \textless 0.05) show notable variations in entropy across meta-information.}
\label{tab:variance_analysis}
\end{table}
\vspace{-10pt}

\section{Results}

\subsection{What meta-information matters}

To quantify the impact of each possible meta-information availability along with the backbone knowledge during adversarial attack construction, we compute the difference between the adversarial attack success rate (ASR) for the target model and the transferability success rate (TSR) from a proxy model, trained from the same backbone, with partial information. We report the results obtained with the PGD attack trained with higher iteration steps per sample as that is more representative for measuring the adversarial attack success in \emph{white-box} and \emph{grey-box} settings. 
%
\paragraph{Which meta-information is important?}
Our results in \cref{fig:whitebox_diff_per_levels} show that the most significant performance decay compared to a \emph{white-box} attack performance occurs when the attacker is unaware of the \emph{mode} of the training of the target model, i.e. if it is trained with complete parameters or only tunes the last classification layers. The second most impactful knowledge for attack construction is the availability of the target tuning \emph{dataset}. The \emph{depth} of the tuning is the least important knowledge for obtaining a transferable attack. We further show in the right part of \cref{fig:whitebox_diff_per_levels} that models that finetune the last classification layers can be trivially exploited with transferable attacks, achieving results significantly better than strong black-box exploitation and closing white-box attack performance. It is, however, apparent that training all of the model weights substantially decreases the efficiency of proxy attacks, with almost no correlation towards meta-information availability.
We further show that our results remain consistent w.r.t. the choice of the dataset, and regardless if the adversarial attack is targeted or untargeted as seen in \cref{fig:whitebox_diff_per_levels_data_target}. It is interesting to note that for datasets with more domain-specific content, such as Oxford-IIIT Pets and Oxford Flowers-102, the effectiveness of the proxy attack dwindles, although these datasets are much less diverse compared to CIFAR-100. 

\paragraph{Meta-information impacts the quality of adversarial attacks}

We also want to measure the effectiveness of the adversarial attack and the impact of meta-information on it by quantifying how the generated adversarial sample has sifted the decision-making of the model. To do this, we compute the entropy of the final softmax layer for each original sample and its adversarial counterpart and complete ANOVA variance analysis \citep{st1989analysis} of entropy distribution. This analysis, presented in \cref{tab:variance_analysis}, tests whether the means of entropies from original and adversarial images differ significantly across the groups of available meta-information. A perfect attack would produce a sample that does not majorly impact the entropy from the model. The analysis reveals that the 
target dataset, and tuning mode significantly influence entropy, particularly in adversarial scenarios. This finding suggests that while this meta-information aids in crafting effective adversarial samples, it also plays a critical role in amplifying entropy shifts, thereby making these adversarial samples more detectable.

To quantify the impact of the meta-information availability during attack construction on the decision-making of the model, we also compute the Jensen-Shannon Divergence \citep{menendez1997jensen} between the output softmax distributions of the model produced for original samples and their adversarial counterparts. High JS divergence suggests a strong attack, as the adversarial example causes a significant shift in the model's predicted probabilities, with minimal changes to the input sample. Our results show that not knowing the \emph{mode} of the target model training causes the most degradation in constructing successful adversarial samples with proxy attacks. The second most important fact is the choice of the target \emph{dataset}, while the \emph{depth} of the final classification layers does not seem to be impactful for creating adversarial samples. This reaffirms our findings from \cref{fig:whitebox_diff_per_levels} and \cref{fig:whitebox_diff_per_levels_tuning}, while also revealing a critical insight: proxy attacks, even when constructed without knowledge of the target model's \emph{dataset} or \emph{depth}, can generate adversarial samples that induce more pronounced distribution shifts than \emph{white-box} attacks. In other words, attackers do not require access to the training dataset or model classification depth to craft adversarial samples capable of significantly disrupting the target model’s decision-making process.

\subsection{Backbone-attacks}

To test the extent of the vulnerabilities that the knowledge of the pre-trained backbone can cause, we evaluate our naive exploitation method, \emph{backbone attack}, that utilizes only the pre-trained feature extractor for adversarial sample construction.
Our results in \cref{fig:whitebox_diff_per_levels} and \cref{fig:whitebox_diff_per_levels_data_target} show that \emph{backbone attacks} are highly effective at producing transferable adversarial samples regardless of the target model tuning \emph{mode}, \emph{dataset} or classification layer \emph{depth}. This naive attack shows significantly higher transferability compared to a strong \emph{black-box} attack with a sizeable query and iteration budget and almost all \emph{proxy attacks}. The results are consistent across all meta-information variations, showing that even a naive attack can exploit the target model vulnerabilities closely to a \emph{white-box} setting, given the knowledge of the pre-trained backbone. Moreover, from \cref{fig:whitebox_diff_per_levels_tuning}, we see that the adversarial samples produced from this attack, on average, cause a bigger shift in the model's decision-making compared to \emph{white-box attacks}. This indicates that backbone attacks amplify the uncertainty in the target model's predictions, making them more disruptive than conventional \emph{white-box} attacks, highlighting the inherent risks of sharing pre-trained backbones for public use. A concerning aspect of backbone attacks is their effectiveness in resource-constrained environments. Unlike black-box attacks, which often require extensive computation or iterative querying, backbone attacks can be executed with minimal resources, leveraging pre-trained models freely available in public repositories. This ease of implementation raises concerns, as it lowers the barrier for malicious actors to exploit adversarial vulnerabilities.

\subsection{Knowing weights vs Knowing everything but the weights}

To isolate the impact of pre-trained backbone knowledge in adversarial transferability, we train two sets of models from the same ResNet-50 SwAV backbone with identical meta-information variations but different batch sizes. This allows the production of two sets of models with matching training meta-information but varying weights; one set is chosen as the target, and the other as the proxy model. We aim to compare the adversarial transferability of the attacks from the set of proxies towards their matching targets with the backbone attacks. This allows us to simulate conditions where adversaries either know all meta-information but lack the weights or have access to the backbone weights alone.
%
\begin{figure}[t!]
    \centering
    \includegraphics[clip=true,width=\columnwidth]{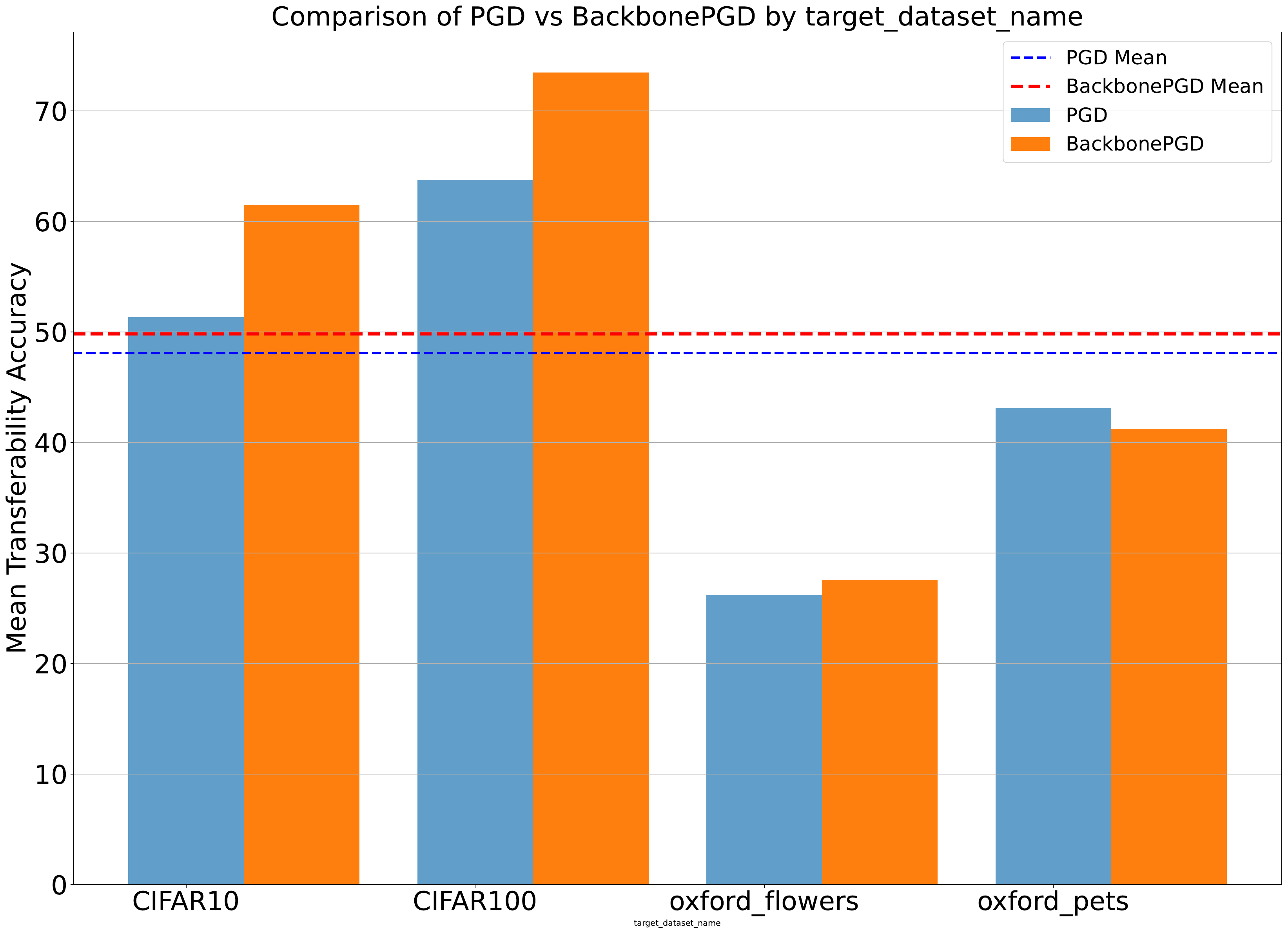}
    \caption{The figure shows scenarios where adversaries either know all meta-information but lack the weights or have access to the backbone weights (SwaV ResNet-50) alone. Knowledge of only the backbone is highlighted as \emph{BackbonePGD}.}
    \label{fig:swav_backbone_meta_info}
\end{figure}
Our results in \cref{fig:swav_backbone_meta_info} show that the knowledge of the pre-trained backbone is, on average, a stronger or at least an equivalent signal for producing adversarially transferable attacks compared to possessing all of the training meta-information without the knowledge of the weights. The results are consistent across all of the datasets, with domain-specific datasets showing marginal differences in adversarial transferability between the two scenarios. This means that possessing information about only the target model backbone is equivalent to knowing all of the training meta-information for constructing transferable adversarial samples.

\section{Conclusions}

In this paper, we investigated the vulnerabilities of machine vision models fine-tuned from publicly available pre-trained backbones under a novel \emph{grey-box} adversarial setting. Through an extensive evaluation framework, including over 20,000 adversarial transferability comparisons, we measured the effect of varying levels of training meta-information availability for constructing transferable adversarial attacks.
We also introduced a naive \emph{backbone attack} method, showing that access to backbone weights is sufficient for obtaining adversarial attacks significantly better than query-based \emph{black-box} settings and approaching white-box performance. We found that attacks crafted using only the backbone weights often induce more substantial shifts in the model's decision-making than traditional white-box attacks. 
We demonstrated that access to backbone weights is equivalent in effectiveness to possessing all meta-information about the target model, making public backbones a critical security concern.
Our results highlight significant security risks associated with sharing pre-trained backbones, as they enable attackers to craft highly effective adversarial samples, even with minimal additional information. These findings underscore the need for stricter practices in sharing and deploying pre-trained backbones to mitigate the inherent vulnerabilities exposed by adversarial transferability.

\section*{Acknowledgments}
$\begin{array}{l}\includegraphics[width=1cm]{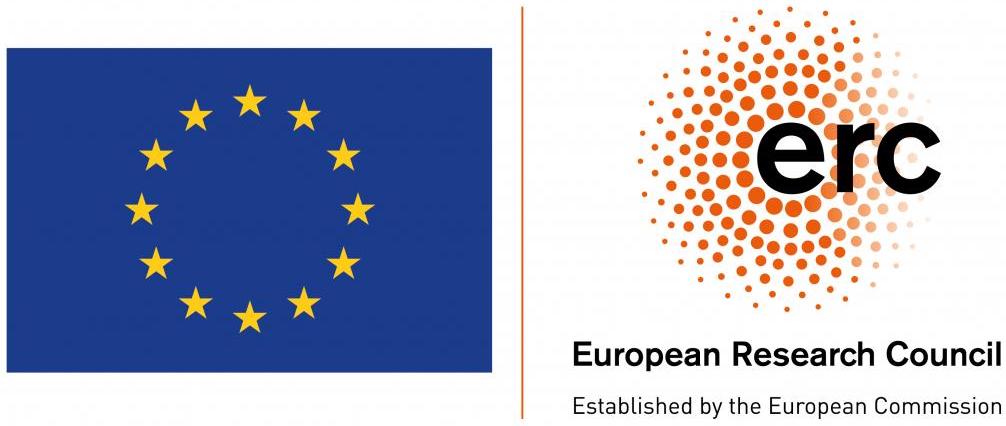} \end{array}$ 
Erik is partially funded by a DFF Sapere Aude research leader grant under grant agreement No 0171-00034B, as well as by an NEC PhD fellowship, and is supported by the Pioneer Centre for AI, DNRF grant number P1.
Pasquale was partially funded by ELIAI (The Edinburgh Laboratory for Integrated Artificial Intelligence), EPSRC (grant no.\ EP/W002876/1), an industry grant from Cisco, and a donation from Accenture LLP.
Isabelle's research is partially funded by the European Union (ERC, ExplainYourself, 101077481), and is supported by the Pioneer Centre for AI, DNRF grant number P1.
This work was supported by the Edinburgh International Data Facility (EIDF) and the Data-Driven Innovation Programme at the University of Edinburgh.

\bibliography{main}
\bibliographystyle{icml2025}

\newpage
\appendix
\onecolumn
\section{Adversarial Transferability per model}

The adversarial transferability for each type of model can be seen in  \cref{tab:transferability_per_model}.

\begin{table}[!t]
\resizebox{\textwidth}{!}{%
\begin{tabular}{@{}lrrrr@{}}
\toprule
\textbf{Model Families} & \textbf{CIFAR10} & \textbf{CIFAR100} & \textbf{Oxford Flowers} & \textbf{Oxford Pets} \\ \midrule
AlexNet (Colorization, IN1K)     & 88.97 & 98.96 & 24.91 & 49.94 \\
AlexNet (Colorization, IN22K)    & 89.56 & 98.92 & 25.19 & 50.06 \\
AlexNet (Colorization, YFCC100M) & 87.84 & 98.55 & 24.91 & 49.96 \\
AlexNet (Jigsaw, IN1K)           & 53.25 & 74.03 & 26.96 & 45.38 \\
AlexNet (Jigsaw, IN22K)          & 53.06 & 73.76 & 30.61 & 49.86 \\
AlexNet (DeepCluster V2)         & 49.59 & 64.38 & 27.15 & 44.52 \\
ResNet-50 (Jigsaw, IN22K)        & 61.03 & 81.81 & 26.37 & 47.28 \\
ResNet-50 (Colorization, IN1K)   & 89.86 & 98.07 & 24.91 & 50.12 \\
ResNet-50 (Colorization, IN22K)  & 88.99 & 97.89 & 27.01 & 50.00 \\
ResNet-50 (Jigsaw, IN1K)         & 56.34 & 80.01 & 25.46 & 48.12 \\
ResNet-50 (Jigsaw, IN22K)        & 54.48 & 75.08 & 26.79 & 47.75 \\
ResNet-50 (RotNet, IN1K)         & 47.71 & 72.61 & 37.86 & 45.69 \\
ResNet-50 (Jigsaw, IN1K)         & 58.02 & 78.32 & 26.17 & 48.06 \\
ResNet-50 (NPID)                 & 58.37 & 80.39 & 49.77 & 48.42 \\
ResNet-50 (PIRL)                 & 58.80 & 84.12 & 34.03 & 44.10 \\
ResNet-101 (SimCLR)              & 55.09 & 70.34 & 28.54 & 47.12 \\
ResNet-50 (SimCLR)               & 51.57 & 65.91 & 30.26 & 44.12 \\
ResNet-50 (SwAV, 400ep)          & 48.63 & 68.46 & 28.79 & 44.33 \\
ResNet-50 (SwAV, 800ep)          & 50.23 & 67.89 & 27.73 & 45.33 \\
DeiT-Small (DINO)                & 63.37 & 85.08 & 26.56 & 47.26 \\
XCiT-Small (DINO)                & 49.46 & 64.84 & 27.19 & 46.76 \\
\bottomrule
\end{tabular}%
}
\caption{Adversarial Transferability Averaged for each dataset per model architecture type}
\label{tab:transferability_per_model}
\end{table}


\end{document}